\documentclass[manuscript,screen,nonacm]{acmart}%

\AtBeginDocument{%
  \providecommand\BibTeX{{%
    \normalfont B\kern-0.5em{\scshape i\kern-0.25em b}\kern-0.8em\TeX}}}

\setcopyright{none}%

\usepackage{tasks}
\usepackage{todonotes}

\usepackage{array}
\newcolumntype{L}[1]{>{\raggedright\let\newline\\\arraybackslash\hspace{0pt}}m{#1}}
\newcolumntype{C}[1]{>{\centering\let\newline\\\arraybackslash\hspace{0pt}}m{#1}}
\newcolumntype{R}[1]{>{\raggedleft\let\newline\\\arraybackslash\hspace{0pt}}m{#1}}

\usepackage{graphicx}
\usepackage{caption}
\usepackage{subcaption}
\usepackage{wrapfig}

\usepackage{booktabs}

\usepackage{enumitem}

\usepackage{fontawesome}

\usepackage[framemethod=tikz]{mdframed}
\definecolor{myblue}{rgb}{0.122, 0.435, 0.698}
\definecolor{mygreen}{rgb}{0.125, 0.525, 0.220}
\definecolor{myyellow}{rgb}{0.588, 0.439, 0.000}
\newmdenv[innerlinewidth=0.5pt,roundcorner=4pt,innerleftmargin=6pt,
          innerrightmargin=6pt,innertopmargin=6pt,innerbottommargin=6pt,
          linecolor=myblue,backgroundcolor=myblue!25!white]{mybluebox}
\newmdenv[innerlinewidth=0.5pt,roundcorner=4pt,innerleftmargin=6pt,
          innerrightmargin=6pt,innertopmargin=6pt,innerbottommargin=6pt,
          linecolor=mygreen,backgroundcolor=mygreen!25!white]{mygreenbox}
\newmdenv[innerlinewidth=0.5pt,roundcorner=4pt,innerleftmargin=6pt,
          innerrightmargin=6pt,innertopmargin=6pt,innerbottommargin=6pt,
          linecolor=myyellow,backgroundcolor=myyellow!25!white]{myyellowbox}

\begin{document}

\title[%
FAT Forensics%
]{%
FAT Forensics:\\ A Python Toolbox for Algorithmic Fairness, Accountability and Transparency%
}

\author{Kacper Sokol}
\email{K.Sokol@bristol.ac.uk}
\email{Kacper.Sokol@rmit.edu.au}
\orcid{0000-0002-9869-5896}
\affiliation{%
  \institution{Intelligent Systems Laboratory, University of Bristol}
  \country{United Kingdom}
}
\affiliation{%
  \institution{ARC Centre of Excellence for Automated Decision-Making and Society, RMIT University}%
  \country{Australia}
}

\author{Raul Santos-Rodriguez}
\email{enrsr@bristol.ac.uk}
\orcid{0000-0001-9576-3905}
\affiliation{%
  \institution{Intelligent Systems Laboratory, University of Bristol}
  \country{United Kingdom}
}

\author{Peter Flach}
\email{Peter.Flach@bristol.ac.uk}
\orcid{0000-0001-6857-5810}
\affiliation{%
  \institution{Intelligent Systems Laboratory, University of Bristol}
  \country{United Kingdom}
}

\renewcommand{\shortauthors}{Sokol, Santos-Rodriguez, and Flach}%

\begin{abstract}
Today, artificial intelligence systems driven by machine learning algorithms can be in a position to take important, and sometimes legally binding, decisions about our everyday lives. %
In many cases, however, these systems and their actions are neither regulated nor certified. %
To help counter the potential harm that such algorithms can cause we developed an open source toolbox that can analyse selected fairness, accountability and transparency aspects of the machine learning process: data (and their features), models and predictions, allowing to automatically and objectively report them to relevant stakeholders. %
In this paper we describe the design, scope, usage and impact of this Python package, which is published under the 3-Clause BSD open source licence.%
\end{abstract}


\maketitle

\begin{mybluebox}
\textbf{Note}\quad%
This is a pre-print of a paper published in the \emph{Software Impacts} journal (\href{https://doi.org/10.1016/j.simpa.2022.100406}{10.1016/j.simpa.2022.100406}).%
\end{mybluebox}
\vspace{1ex}

\begin{myyellowbox}
\textbf{Highlights}%
\begin{itemize}[topsep=0pt,label=\faLightbulbO,leftmargin=.5cm,itemindent=0.0cm,labelwidth=0.5cm,labelsep=0cm,align=left]%
  \item A Python toolbox for algorithmic Fairness, Accountability and Transparency (FAT).%
  \item Built atop SciPy and NumPy, and distributed under the 3-Clause BSD licence (new BSD).
  \item Based on a modular architecture that allows to compose bespoke FAT tools with ease.%
  \item Supports research and deployment modes of operation that enable diverse use cases.%
  \item Accompanied by comprehensive documentation, examples, tutorials and how-to guides.
\end{itemize}
\end{myyellowbox}
\vspace{1ex}

\begin{mygreenbox}
\textbf{Code Metadata}\\[.15cm]%
\begin{tabular}{p{3.8cm}p{10.1cm}}
\toprule
Software version & 0.1.1 \\
Code repository & \url{https://github.com/fat-forensics/fat-forensics/} \\
Reproducible capsule & \url{https://codeocean.com/capsule/8437308/tree/v1/} \\
Licence & 3-Clause BSD Licence (New BSD) \\
Versioning control system & git \\
Programming language & Python \\
Requirements \& dependencies & \url{https://fat-forensics.org/getting_started/install_deps_os.html#installation-instructions} \\
Developer documentation & \url{https://fat-forensics.org/}\\
Contact details & \url{https://fat-forensics.org/#communication} \\
\bottomrule
\end{tabular}
\end{mygreenbox}

\section{Algorithmic Fairness, Accountability and Transparency with FAT Forensics\label{sec:intro}}%

Open source software is the backbone of reproducible research, especially so in Artificial Intelligence (AI) and Machine Learning (ML) where changing the seed of a random number generator may cause a state-of-the-art solution to become a subpar predictive system. %
Despite numerous efforts to ensure that publications are accompanied by code, both the AI and ML fields struggle with a reproducibility crisis~\cite{hutson2018artificial}. %
One way to address this problem is to promote publishing high-quality software used for scientific experiments under an open source licence or enforce it as part of the publishing process~\cite{sonnenburg2007need}. %
In spite of their importance, implementations are nonetheless commonly treated just as a research by-product and often abandoned after publishing the findings based upon them. %
We call this phenomenon \emph{paperware}, i.e., code whose main purpose is to see a paper towards publication rather than implement any particular concept with thorough software engineering practice. %
Such attitude results in standalone packages that often prove difficult to use due to the lack of documentation, testing, usage examples and (post-publication) maintenance, therefore impacting their reach, usability and, more broadly, reproducibility of scientific findings. %
This state of affairs is especially problematic for AI and ML research with its fast-paced environment, lack of standards and far-ranging social implications.%

Widespread reliability issues with ML systems have inspired a range of frameworks to assess and document them as well as report their quality, robustness and other (technical) properties through standardised mechanisms. %
For example, researchers have suggested approaches to characterise data sets~\cite{gebru2018datasheets,holland2018dataset}; automated decision-making systems~\cite{reisman2018algorithmic}; predictive models offered as a service accessible via an Application Programming Interface (API)~\cite{hind2018increasing}; ranking algorithms~\cite{yang2018nutritional}; AI \& ML explainability approaches~\cite{sokol2020explainability}; and privacy aspects of applications that collect, process and share user data~\cite{kelley2009nutrition} to ensure their high quality, transparency, reliability and accountability. %
Such efforts are laudable, however they may require the authors to understand the investigated system in detail, suffer from limited scope or be subject to time- and labour-intensive creation process, all of which can hinder their uptake or slow down the ML research and development cycle. %
Moreover, self-reporting -- and lack of external audits -- means that some of their aspects may be subjective, hence misrepresent the true behaviour of the underlying system, whether done intentionally or not. %
Certification, on the other hand, creates a need for external bodies, which seems difficult to achieve for all ML systems that somehow affect humans.%

To help address such shortcomings in the fields of AI \& ML Fairness, Accountability and Transparency (FAT), we designed and developed an open source Python package called \texttt{FAT Forensics}~\cite{sokol2020fat} -- Table~\ref{tab:implementation} lists the algorithms distributed in its latest release (version 0.1.1). %
It is intended as an interoperable framework to \emph{implement}, \emph{test} and \emph{deploy} novel algorithms proposed by the FAT community as well as facilitate their evaluation and comparison against state-of-the-art methods, therefore democratising access to these techniques. %
The toolbox is capable of analysing all facets of the data-driven predictive process -- data (raw and their features), models and predictions -- in view of their FAT aspects. %
The common interface layer of the software (described in \S\ref{sec:design_sec}) makes it flexible enough to support workflows typical of academics and practitioners alike, and enables two \emph{modes of operation} -- \emph{research} and \emph{deployment} -- that span diverse use cases such as prototyping, exploratory analytics, (numerical or visual) reporting and dashboarding as well as inspection, monitoring and evaluation of FAT properties. %
Additionally, the package is backed by thorough and beginner-friendly documentation, which spans tutorials, examples, how-to manuals and a user guide. %
In the following section (\S\ref{sec:design_sec}) we introduce our software and describe its architecture. %
Next, we present a number of possible use cases and benefits of having various FAT algorithms under a shared roof (\S\ref{sec:examples}). %
We conclude the paper with an overview of the impact of our package to date and a discussion of the envisaged long-term benefits of \texttt{FAT Forensics} in view of our contributions (\S\ref{sec:impact}). %
While this paper focuses on the wide-reaching advantages of our software, a complementary publication~\cite{sokol2020fat} offers its high-level overview, implementation details and comparison to related packages.%

\begin{table}[t]
\centering
\footnotesize
\begin{tabular}{L{1.3cm}L{4.0cm}L{4.0cm}L{4.0cm}}
\toprule
 & \textbf{Fairness} & \textbf{Accountability} & \textbf{Transparency}\\
\midrule
  \textbf{Data \& Features} &
  \(\bullet\)~Systemic Bias\newline\(\bullet\)~Sub-population Representation &%
  \(\bullet\)~Sampling Bias\newline\(\bullet\)~Data Density Checker &%
  \(\bullet\)~Data Description\newline\(\bullet\)~Summary Statistics\\
\midrule
  \textbf{Models} &
  \(\bullet\)~Group-based Fairness &%
  \(\bullet\)~Group-based Performance Metrics\newline\(\bullet\)~Systematic Performance Bias &%
  \(\bullet\)~Global Surrogates (bLIMEy)\newline\(\bullet\)~Partial Dependence\newline\(\bullet\)~Submodular Pick\\
\midrule
  \textbf{Predictions} &
  \(\bullet\)~Counterfactual Fairness &%
  \(\bullet\)~Prediction Confidence &%
  \(\bullet\)~Model-agnostic Counterfactuals\newline\(\bullet\)~Local Surrogates (bLIMEy)\newline\(\bullet\)~LIME (bLIMEy implementation)\newline\(\bullet\)~Individual Conditional Expectation\\
\bottomrule
\end{tabular}
\caption{FAT functionality implemented in the latest release -- version 0.1.1 -- of \texttt{FAT Forensics}.\label{tab:implementation}}%
\end{table}

\section{Design and Architecture\label{sec:design_sec}}%
Systematic evaluation and comparison of AI \& ML techniques is an active area of research across many different communities. %
In well-established research fields, such as supervised learning, we can observe convergence towards commonly accepted (predictive) performance metrics and evaluation software; %
their implementations often constitute a fundamental part of relevant packages, nonetheless the independence of many such metrics from the underlying predictive algorithms allows for standalone software dedicated to calculating them, e.g., \emph{PyCM}~\cite{haghighi2018}. %
In contrast, relatively young fields -- such as algorithmic fairness, accountability (robustness, safety, security \& privacy) and transparency (interpretability \& explainability) -- usually lack this type of evaluation strategies and software solutions, making them a welcome addition that has the potential to streamline research.%

To address these challenges, we developed an open source Python framework for evaluating, comparing and deploying FAT algorithms. %
We chose Python because of its prevalence across different AI \& ML research communities and overall simplicity. %
We opted for a minimal (required) dependency on NumPy and SciPy to facilitate easy deployment in a variety of settings. %
An optional dependency on Matplotlib, scikit-learn, Pillow and scikit-image %
enables access to basic visualisations, ML algorithms and image manipulations (needed by explainability functions). %
The toolbox is hosted on GitHub %
to facilitate community contributions, and released under the 3-Clause BSD licence to open it up for commercial applications. %
To encourage long-term sustainability it has been developed in accordance with the best software engineering practices such as: %
unit and integration testing; %
high code coverage; %
continuous integration; %
function- and module-level technical API documentation; %
task-focused code examples; %
narrative-driven tutorials; %
problem-oriented how-to guides; and %
a comprehensive user guide. %
The toolbox implements a number of popular FAT algorithms -- with many more to come -- under a coherent API, reusing many functional components across FAT tools and making them readily accessible to the community. %
The initial development is focused on tabular data and well-established predictive models (scikit-learn~\cite{pedregosa2011scikit-learn}), which will be followed by techniques capable of handling sensory data (images \& text) and neural networks (TensorFlow~\cite{abadi2016tensorflow} \& PyTorch~\cite{paszke2019pytorch}). %
Additionally, we envisage that relevant software packages that are already prominent in the FAT community and that adhere to best software engineering practice can be ``wrapped'' by our toolbox under a common API to make them easily accessible and avoid re-implementing them.%

Algorithms included in \texttt{FAT Forensics} are designed and engineered to support two main application areas. %
The \emph{research mode}, characterised by ``data in -- visualisations out'', envisages the toolbox being loaded into an interactive Python session (e.g., a Jupyter Notebook) to support exploratory analysis, prototyping, development, evaluation and testing. %
This mode is intended for researchers who could use it to propose new fairness metrics, compare them with existing solutions or inspect a new predictive system or data set (without the burden of setting up a dedicated software engineering workflow). %
Contributing these implementations of cutting-edge techniques to \texttt{FAT Forensics} will in turn make the package attractive for monitoring and auditing of data-driven systems -- the second intended application domain. %
More specifically, the \emph{deployment mode}, characterised by ``data in -- data out'', offers to incorporate the package into a data processing pipeline to provide a (numerical) analytics, hence support any kind of automated reporting, dashboarding or certification (thus partially alleviating the issues with manual, error-prone and subjective characterisation of AI \& ML components). %
This mode is intended for ML practitioners who (by accessing the low-level API) may use it to monitor or evaluate a data-driven system; where continuous integration is used in software engineering to ensure high quality of the code, our toolbox could be employed to evaluate FAT of any component of an ML pipeline during its development and deployment.%

A considerable portion of FAT software is developed to support research outputs, %
which often results in superfluous dependencies, data sets, predictive models and (interactive) visualisations being distributed with the code base that itself is accessible via a non-standard API. %
To mitigate these issues, \texttt{FAT Forensics} decouples the core FAT functionality from its possible presentation to the user and experiment-specific resources. %
This abstraction of the software infrastructure is achieved by making minimal assumptions about the operational setting of these algorithms, therefore facilitating a common interface layer for key FAT functionality, focusing only on the interactions between data, models, predictions and users~\cite{sokol2020fat}. %
In this purview a predictive model is assumed to be a plain Python object with \texttt{fit}, \texttt{predict} and, optionally, \texttt{predict\_proba} methods, which offers compatibility with scikit-learn~\cite{pedregosa2011scikit-learn} -- the most popular Python ML toolbox -- without explicitly depending on it, in addition to supporting %
any other predictor that can be represented in this way, e.g., TensorFlow, PyTorch or even one hosted on the Internet and accessible via a web API. %
Similarly, a data set is assumed to be a two-dimensional NumPy array: either a classic or a structured array, with the latter bringing support for (string-based) categorical attributes. %
Since visualisations are a vital part of our first application mode (research), the software provides basic plotting functionality that is only enabled when the \emph{optional} Matplotlib dependency is installed. %
In addition to relaxed input requirements, all of the techniques incorporated into the package are split into interoperable algorithmic building blocks that can be easily reused, even across FAT borders, to create new functionality -- the versatility of this atomic-level decomposition is demonstrated in the following section. %
More details about the technical aspects of the software can be found in the \texttt{FAT Forensics} technical paper~\cite{sokol2020fat}.%

\section{Use Cases\label{sec:examples}}%
We present three distinct use cases to demonstrate how the software can be applied to analyse FAT aspects of real data, illustrating the diverse range of functionality enabled by its universal infrastructure. %
To this end, we employ the UCI Census Income (Adult) data set~\cite{kohavi1996census}, which is popular in algorithmic fairness and transparency research. %
The data analysis that follows is representative of the \emph{research mode} and is inspired by the tutorials included in the \texttt{FAT Forensics} documentation%
\footnote{\url{https://fat-forensics.org/tutorials/index.html}}; %
it can be reproduced with a dedicated Jupyter Notebook%
\footnote{\url{https://github.com/fat-forensics/resources/blob/master/fat_forensics_overview/FAT_Forensics.ipynb}}. %
To demonstrate the \emph{deployment mode}, we provide a dashboard based on Plotly Dash, which facilitates interactive analysis of the same data set using \texttt{FAT Forensics} as the back end\footnote{\url{https://fatf.herokuapp.com/}. (Source code available at: \url{https://github.com/fat-forensics/fatf-dashboard/}.)}.%

\begin{figure}[t]
\begin{minipage}[t]{.59\textwidth}
    \centering
    \begin{subfigure}[b]{0.315\textwidth}%
        \centering
        \includegraphics[width=\textwidth]{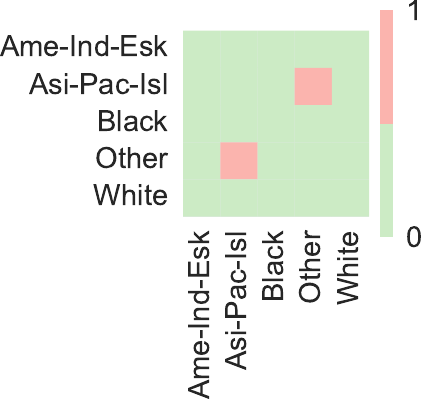}%
       \caption{Equal accuracy.\label{fig:adult_fair_race:acc}}%
    \end{subfigure}
    \hspace{0.01\textwidth}%
    \begin{subfigure}[b]{0.315\textwidth}%
        \centering
        \includegraphics[width=\textwidth]{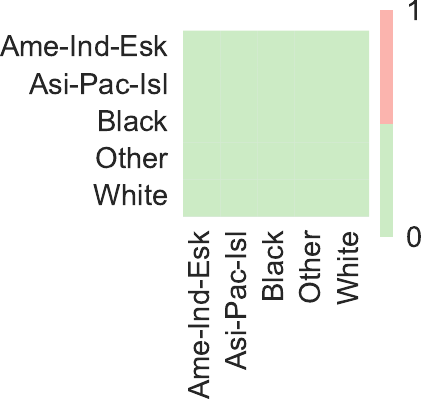}%
       \caption{Equal opportunity.}%
    \end{subfigure}
    \hspace{0.01\textwidth}%
    \begin{subfigure}[b]{0.315\textwidth}%
        \centering
        \includegraphics[width=\textwidth]{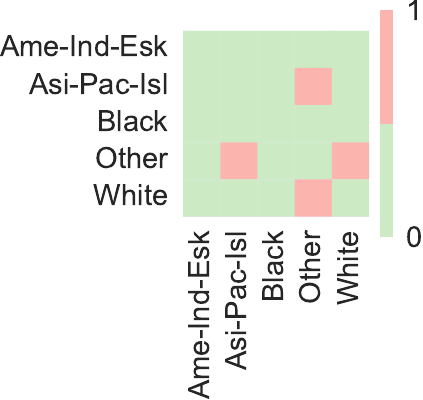}%
       \caption{Demographic parity.}%
    \end{subfigure}
    \caption{Pairwise group-based \textbf{fairness} for the \emph{race} feature of the Adult data set. Red (1) indicates disparate impact for a given pair of sub-populations and green (0) conveys that they are treated comparably.\label{fig:adult_fair_race}}%
\end{minipage}
\hfill
\begin{minipage}[t]{.39\textwidth}
    \centering
    \begin{subfigure}[b]{0.475\textwidth}%
        \centering
        \includegraphics[width=1.0\textwidth]{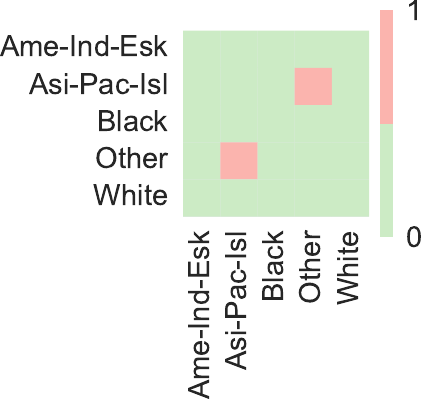}%
       \caption{Predictive accuracy.\label{fig:adult_perf_race:acc}}%
    \end{subfigure}
    \hspace{0.01\textwidth}
    \begin{subfigure}[b]{0.475\textwidth}%
        \centering
        \includegraphics[width=1.0\textwidth]{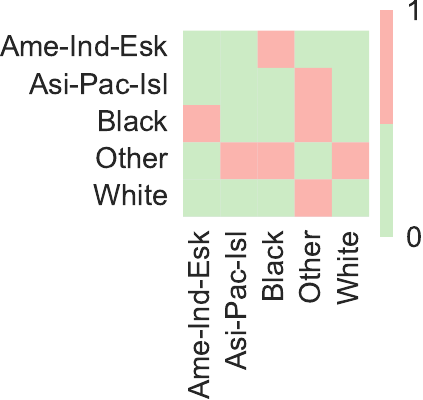}%
       \caption{True negative rate.\label{fig:adult_perf_race:tfn}}%
    \end{subfigure}
    \caption{%
Pairwise group-based \textbf{performance} disparity for the \emph{race} feature of Adult. %
Red (1) shows disparate performance and green (0) comparable treatment.%
\label{fig:adult_perf_race}}%
\end{minipage}
\end{figure}

\paragraph{Feature Grouping}%
One of the core building blocks of \texttt{FAT Forensics} is a collection of functions to partition data based on (sets of) unique values %
for categorical features and threshold-based binning for numerical attributes. %
This algorithmic concept -- in conjunction with any standard (predictive) performance metric derived from predicted and true labels -- facilitates a number of FAT workflows. %
A variety of different group-based (pairwise) \textbf{fairness} criteria, not limited to the ones implemented in the package, can be computed in this way by conditioning on protected features (attributes that may be used for discriminatory treatment, e.g., gender), allowing us to investigate disparate impact of a predictive model based on group unaware, equal opportunity, equal accuracy or demographic parity metrics, among many others~\cite{hardt2016equality}. %
Since some of them are mutually incompatible~\cite{miconi2017note}, comparing them side-by-side can be beneficial. %
For example, the \emph{Asian-Pac-Islander} (\emph{Asi-Pac-Isl}) and \emph{Other} groups are subject to fairness disparity when \emph{equal accuracy} and \emph{demographic parity} are considered;
\emph{Other} and \emph{White} sub-populations are also treated unfairly according to \emph{demographic parity}; %
whereas \emph{equal opportunity} does not exhibit any signs of disparate impact %
as shown in Figure~\ref{fig:adult_fair_race}.%

\begin{figure}[t]
    \centering
    \begin{subfigure}[b]{0.18\textwidth}%
        \centering
        \includegraphics[height=4.0cm]{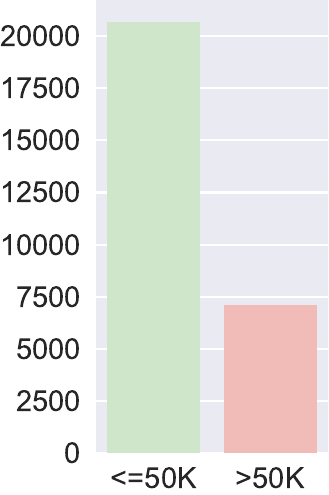}%
       \caption{White.}
    \end{subfigure}
    \hspace{0.01\textwidth}
    \begin{subfigure}[b]{0.18\textwidth}%
        \centering
        \includegraphics[height=4.0cm]{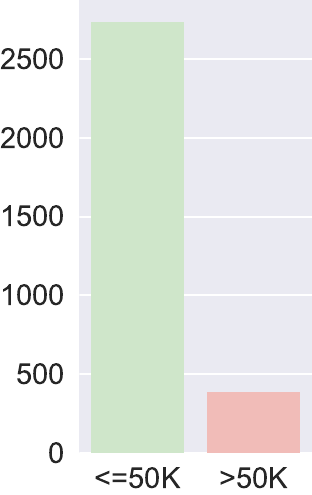}%
       \caption{Black.}
    \end{subfigure}
    \hspace{0.01\textwidth}
    \begin{subfigure}[b]{0.18\textwidth}%
        \centering
        \includegraphics[height=4.0cm]{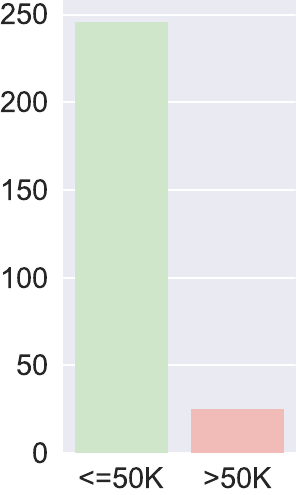}%
       \caption{Other.}
    \end{subfigure}
    \hspace{0.01\textwidth}
    \begin{subfigure}[b]{0.18\textwidth}%
        \centering
        \includegraphics[height=4.0cm]{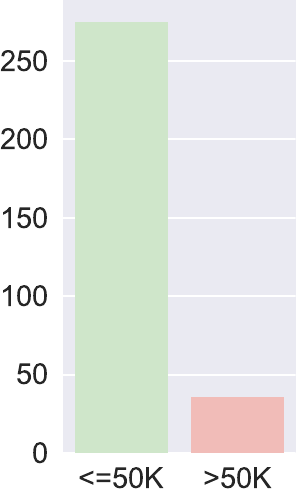}%
       \caption{Ame-Ind-Esk.}
    \end{subfigure}
    \hspace{0.01\textwidth}
    \begin{subfigure}[b]{0.18\textwidth}%
        \centering
        \includegraphics[height=4.0cm]{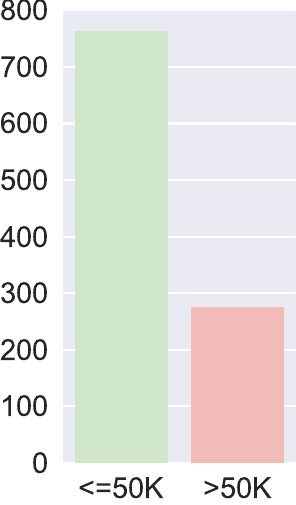}%
       \caption{Asi-Pac-Isl.}
    \end{subfigure}
    \caption{Income distribution for each unique value of the \emph{race} feature in the Adult data set.\label{fig:adult_data_race}}
\end{figure}

The grouping functionality can also help to assess \textbf{accountability} of data and models in a similar fashion. %
For example, sample-size disparity across sub-populations in a given data set may cause a \emph{systematic bias} in predictive performance of the resulting model over such (protected) groups since it is likely to under-perform for under-represented individuals. %
This effect can be observed when splitting Adult based on the \emph{race} feature while measuring \emph{accuracy} and \emph{true negative rate}. %
As expected, the former (Figure~\ref{fig:adult_perf_race:acc}) provides the same result as group-based fairness analysis under \emph{equal accuracy} (Figure~\ref{fig:adult_fair_race:acc}); the latter (Figure~\ref{fig:adult_perf_race:tfn}), on the other hand, reveals that four sub-population pairs exhibit significant performance differences, with \emph{Other} affected the most by diverging from all the other groups except \emph{Amer-Indian-Eskimo} (\emph{Ame-Ind-Esk}). %
Partitioning is also useful for \textbf{transparency} analysis; for example, summary statistics such as the distribution of labels across sub-populations based on (protected) features can be generated for a data set prior to modelling to uncover any class imbalance. %
Studying the \emph{race} attribute in this context -- Figure~\ref{fig:adult_data_race} -- reveals that while the classes are skewed across all the splits, the strongest disproportion affects the \emph{Other}, \emph{Ame-Ind-Esk} and \emph{Black} sub-populations.%

\paragraph{Data Density}%
Density estimate for a region in which a data point of interest is located (based on the distribution of training data) can be treated as a proxy for the confidence of its prediction~\cite{perello2016background}, thus helping to judge its \textbf{accountability} and robustness as dense regions should offer more accurate modelling. %
To this end, \texttt{FAT Forensics} implements a bespoke neighbour-based density estimator %
-- its scores are between 0 and 1, where high values are assigned to instances from sparse regions since their n\textsuperscript{th} neighbour (a user-defined parameter) is relatively distant. %
As an illustration we estimate the density of Adult based on its first 1,000 instances and select four data points -- two from a dense and two from a sparse region -- to assess robustness of their predictions. %
The former two receive density scores of 0 and are correctly predicted as \emph{\(\leq\)50K}; %
the latter two are assigned density scores of 1 with one predicted correctly and the other misclassified as \emph{\(\leq\)50K}. %
Upon closer inspection this data point has a relatively high value (99.99\textsuperscript{th} percentile) of the \emph{fnlwgt} feature (1,226,583), which is a clue to its high density score and incorrect prediction (see the aforementioned Jupyter Notebook for more details).%

In addition to engendering trust in predictions, a density estimate can help to assess the quality of exemplar explanations and compute realistic counterfactuals~\cite{poyiadzi2020face}, which can be used as a \textbf{transparency} tool and individual \textbf{fairness} mechanism (by conditioning on protected attributes). %
Sourcing counterfactuals from sparse regions may yield explanations based on instances that are unlikely to occur in the real life, e.g., prescribing a person to become 200 years old. %
Explaining the aforementioned misclassified data point taken from a sparse region provides explanations such as: %
(i)~raising \emph{capital-gain} from 0 to 25,000 predicts \emph{>50K} (sparse region with 1 density score); and %
(ii)~increasing \emph{capital-loss} from 0 to 4,000 and decreasing \emph{fnlwgt} from 1,226,583 to 430,985 predicts \emph{>50K} (dense region with 0.02 density score). %
While (i) prescribes a sensible action, preserving the unusually high value of \emph{fnlwgt} makes it unlikely; %
(ii), on the other hand, decreases the value of this attribute -- therefore placing the counterfactual in a dense region -- and shows that even with 4,000 of \emph{capital-loss} being classified as \emph{>50K} is possible, casting even more suspicion on the unusually high original value of the former feature. %
Finally, no counterfactuals conditioned on protected attributes could be found for this instance, showing us that its prediction is fair (again, see the aforementioned Jupyter Notebook for more details).%

\paragraph{Surrogate Modularity}%
Surrogate explainers are a popular interpretability technique that fits a transparent model in a selected neighbourhood to approximate and explain the predictive behaviour of the underlying black %
box in said region~\cite{craven1996extracting,ribeiro2016why,sokol2019blimey}. %
Given their high modularity, \texttt{FAT Forensics} implements their core building blocks via the bLIMEy meta-algorithm\footnote{\url{https://fat-forensics.org/how_to/transparency/tabular-surrogates.html}} -- consisting of interpretable representation composition, data sampling and explanation generation steps -- which allows the user to easily construct a bespoke surrogate that is suitable for the problem at hand, thus considerably improving the quality and faithfulness of the resulting explanations~\cite{sokol2019blimey,sokol2021towards}. %
For example, an interpretable %
representation of tabular data %
can be built with quartile-%
based discretisation or a feature space partition extracted from a decision tree (the latter is more %
faithful~\cite{sokol2020towards}); %
data can be augmented with Gaussian or mixup~\cite{zhang2018mixup} sampling (the latter offers a diverse and local sample~\cite{sokol2019blimey}); and an explanation can be generated with a linear model or a decision tree %
\begin{wrapfigure}[12]{r}{0.5\textwidth}%
    \centering
    \begin{subfigure}[b]{0.225\textwidth}%
        \centering
        \includegraphics[width=1\textwidth]{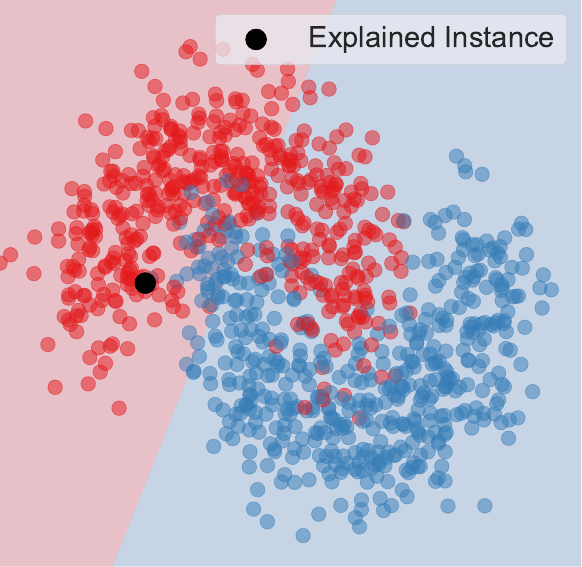}%
        \caption{Linear surrogate.\label{fig:blimey_linear}}%
    \end{subfigure}
    \hspace{0.01\textwidth}
    \begin{subfigure}[b]{0.225\textwidth}%
        \centering
        \includegraphics[width=1\textwidth]{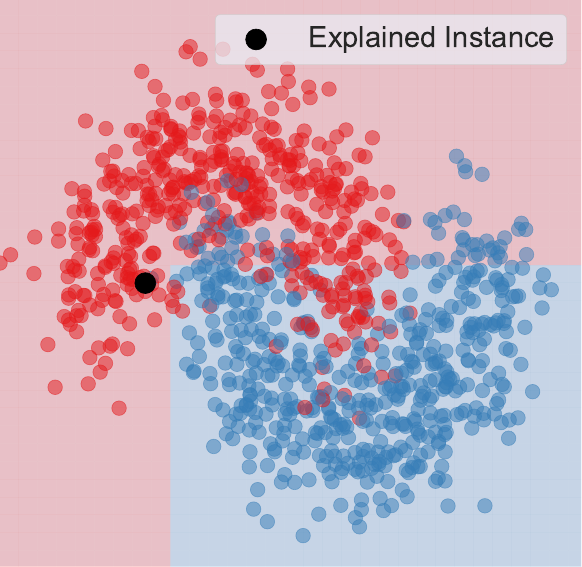}%
        \caption{Tree-based surrogate.\label{fig:blimey_tree}}%
    \end{subfigure}
    \caption{%
Examples of (\subref{fig:blimey_linear}) linear (LIME-like) and (\subref{fig:blimey_tree}) tree-based local surrogates without an interpretable representation (enabling direct visualisation) built with the bLIMEy framework for the Two Moons data set.%
\label{fig:surrogates}}
\end{wrapfigure}
(the former is limited to feature influence, whereas the latter provides a diverse range of insights such as rules and counterfactuals~\cite{sokol2020limetree,sokol2020towards,sokol2021towards}). %
Such a surrogate explainer can either be local -- by sampling data in the neighbourhood of a selected instance -- or global -- when the sample covers the entire data space. %
Specifically, consider the two local surrogates shown in Figure~\ref{fig:surrogates}, where a tree-based explainer~\cite{sokol2020limetree,sokol2020towards} is better able to approximate the decision boundary close to the selected instance.%

\vspace{-.2\baselineskip}
\section{Impact Overview\label{sec:impact}}

While software is one of the primary drivers of progress in AI \& ML research, its quality is often found lacking. %
\texttt{FAT Forensics} offers a possible solution in the space of algorithmic fairness, accountability and transparency by facilitating the development, evaluation, comparison and deployment of FAT tools. %
Sharing a common functional base between implementations of FAT algorithms is one of many advantages of such a comprehensive package. %
Its versatility as well as support for the \emph{research} and \emph{deployment} operation modes make it appealing to members of academia and industry, especially as it supports investigating FAT aspects of an entire predictive pipeline: data, models and predictions. %
This in turn ought to encourage the community to adopt the software and contribute their novel algorithms and bug fixes here (instead of releasing them as standalone code), thus exposing them to the wider audience in a robust and sustainable environment, enhancing reproducibility of research in this space and orienting the package towards real-world use cases. %
By developing FAT tools on a modular level from the ground up \texttt{FAT Forensics} ensures their robustness and accountability in addition to being shielded from any errors that otherwise could have been introduced downstream. %
For example, LIME~\cite{ribeiro2016why} -- which is ``wrapped'' by Microsoft's Interpret~\cite{nori2019interpretml} and Oracle's Skater~\cite{kramer2018skater} libraries -- has known issues with the locality and coherence of its explanations~\cite{laugel2018defining,sokol2019blimey}, which inadvertently affect both these packages. %
We therefore hope and expect that all the software engineering best practice followed during the initial development of \texttt{FAT Forensics} (and maintained carrying forwards) have helped us to create a sustainable package that is easy to extend and contribute to, serving the community for a long time to come.%

Additionally, the modular design of the package facilitates conducting cutting-edge research. %
To date, the implementation of surrogate explainers available in \texttt{FAT Forensics} allowed us to carefully study their capabilities and failure modes, leading to new findings, theories and transparency tools. %
bLIMEy -- the surrogate meta-algorithm -- is a case in point; its inception was inspired by identifying independent algorithmic modules, whose further investigation showed the importance of local sampling for tabular data and effectiveness of decision trees as surrogate models~\cite{sokol2019blimey,sokol2021towards}. %
One particular realisation of this explainer -- LIMEtree -- is based on multi-output regression trees and improves upon many shortcomings of surrogates by offering faithful, consistent, customisable and multi-class explanations of different types, including counterfactuals~\cite{sokol2020limetree}. %
Diverse implementations of surrogate building blocks also helped us to analyse the role and parameterisation of interpretable representations and improve their robustness -- they translate the low-level data representation used by predictive models into human-comprehensible concepts underlying explanations and are the backbone of surrogates~\cite{sokol2020towards}. %
\texttt{FAT Forensics} has also been the foundation of a hands-on conference tutorial on ML explainability~\cite{sokol2020tut} as well as numerous lectures, summer school sessions, educational events and learning resources\footnote{\url{https://events.fat-forensics.org/}}.%

\renewcommand\acksname{Acknowledgements}
\begin{acks}
This work was financially supported by Thales, and is the result of a collaborative research agreement between Thales and the University of Bristol. %
KS and PF were partially supported by TAILOR (Trustworthy AI through Integrating Learning, Optimisation and Reasoning), a project funded by EU Horizon 2020 research and innovation programme under GA No 952215. %
Additionally, KS is supported by the ARC Centre of Excellence for Automated Decision-Making and Society, funded by the Australian Government through the Australian Research Council (project number CE200100005); and %
RSR is supported by the UKRI Turing AI Fellowship EP/V024817/1. %
The authors would also like to acknowledge contributions of student software engineers: Alexander Hepburn, Rafael Poyiadzi and Matthew Clifford.%
\end{acks}

\bibliographystyle{ACM-Reference-Format}
\bibliography{si}

\end{document}